\documentclass[11pt]{article}
\usepackage{coling2018}
\usepackage{times}
\usepackage{url}
\usepackage{latexsym}
\usepackage{multirow}
\usepackage{graphicx}

\RequirePackage{color}

\title{Translating Natural Language Instructions for Behavioral Robot Navigation with a Multi-Head Attention Mechanism}

\author{Patricio Cerda-Mardini, Vladimir Araujo, Alvaro Soto\\
Pontificia Universidad Catolica de Chile \\
Millennium Institute for Foundational Research on Data \\
{\tt \{pcerdam, vgaraujo\}@uc.cl, asoto@ing.puc.cl} \\}

\date{}

\begin{document}
\maketitle
\begin{abstract}
  We propose a multi-head attention mechanism as a blending layer in a neural network model that translates natural language to a high level behavioral language for indoor robot navigation. We follow the framework established by \cite{zang-etal-2018-translating} that proposes the use of a navigation graph as a knowledge base for the task. Our results show significant performance gains when translating instructions on previously unseen environments, therefore, improving the generalization capabilities of the model.
\end{abstract}

\vspace{0.2cm}

\section{Background}
\label{intro}
\begin{figure}[h]
\centering
\label{unique}
\caption{(a) Map of an environment. (b) Its behavioral navigation graph. (c) Proposed model. The natural language instruction on (c) is translated to a sequential behavior plan. The path in (a) and the node-edges in (b), both highlighted in red, correspond to the behaviors predicted by the model (c).}
\includegraphics[width=\textwidth]{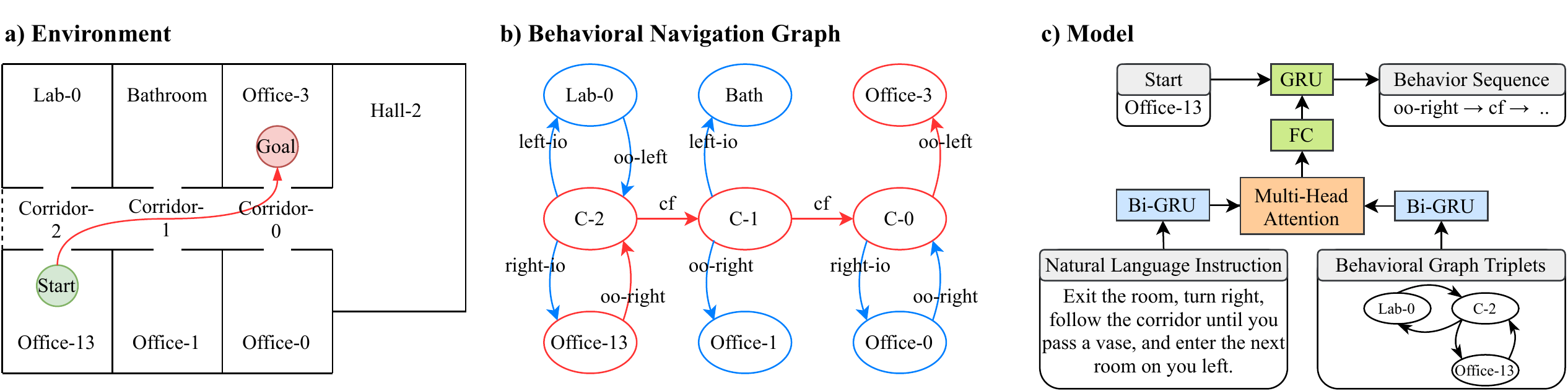}
\end{figure}

Developing robotic agents that can follow natural language instructions remains an open challenge. Ideally, a robot should be able to correctly create an executable navigation plan given a natural language instruction by a user. The objective is to reach a destination from a starting point in a complex but known indoor environment (Figure~\ref{unique}(a)), which could be represented as a graph \cite{sepulveda2018deep}, where the nodes correspond to locations (e.g., office, bedroom), and the edges represent high-level behaviors (e.g., follow corridor, exit office) that allow a robot to navigate between neighboring nodes (Figure~\ref{unique}(b)). We assume the robot can robustly execute every high level behavior, as in \cite{sepulveda2018deep}.

Previous works pose this problem as a translation of instructions to a plan of sequentially executed high-level behaviors \cite{Zang2018BehavioralIN}, leveraging the environment topology through its graph representation \cite{zang-etal-2018-translating}. Specifically, a supervised learning model takes as input a text instruction from the user, the robot initial location, and the behavior graph of the environment encoded as triplets $(n_1, b, n_2)$, where $n_1, n_2$ are places and $b$ the behavior that connects both. It then predicts a sequence of behaviors to reach the instructed destination by means of a typical sequence-to-sequence model with a single soft attention layer that fuses the graph and instruction information. However, at inference time this approach suffers a severe performance hit on environments that were not seen during training. In this work, we propose to modify the attention layer by using a multi-headed mechanism that improves the model generalization capabilities, therefore, increasing performance in unseen environments.

\section{Methodology}

\paragraph{\textbf{Approach}} Inspired by the success of the Transformer model \cite{vaswani} on encoding different relationships of multi-modal data \cite{tan2019lxmert,zhou2020recommending}, we propose to use its multi-head attention mechanism to blend information from the two representation sub-spaces, natural instructions and navigation graph, in a more useful way. That is, different heads will specialize in fusing different patterns between both information sources. We hypothesize that this capability might help the decoder to alleviate the performance hit in novel environments at test time.

\paragraph{\textbf{Proposed Model}} The architecture (Figure~\ref{unique}(c)) considers an initial encoding layer, where each word of the instruction is encoded using pre-trained GloVe descriptors \cite{Glove}, and each triplet set is one-hot encoded to indicate which of the $\mathcal{B}$ behaviors and $\mathcal{N}$ nodes constitute each triplet. Subsequently, the encodings are embedded using bi-directional Gated Recurrent Units (GRU) \cite{GRU}. The multi-modal representations are then fused by the newly added multi-head attention mechanism. A fully connected layer downstream reduces the dimensionality of the fused information, which is used as context $C$ by a recurrent GRU decoder. The decoder takes the initial position and translates the instruction to a sequential behavioral plan, soft attending its context $C$ at each time step. The loss function is cross entropy with respect to correct translations.

\paragraph{\textbf{Experimental Setup}} We use the dataset introduced in \cite{zang-etal-2018-translating} with the original train and test splits, where the Test-Repeated split has environments that were seen by the agent at training time, and the Test-New split has previously unseen maps. In total, we consider 10,040 instructions (8,066 for training) distributed across 100 maps, each with 6 to 65 rooms. We also use the same performance metrics: F1 score, edit distance (ED) to ground truth, and M@k metrics, where we have a match if the translation is under $k$ moves away from the ground truth\footnote{A move adds, deletes, or swaps a behavior in the plan.}, with M@0 being an exact match. The model was trained for 200 epochs, with a batch size of 256. The multi-head attention layer was set to have 4 heads. The rest of the model parameters are as established in \cite{zang-etal-2018-translating}.

\section{Results \& Discussion}
\paragraph{\textbf{Results}} Table~1 details the performance of our approach, along with the baseline reported by \cite{zang-etal-2018-translating} as well as by our own implementation of that model, which notably was not able to perform as expected on the Test-Repeated set. As a result of using our multi-headed approach, we see a clear performance gain (23.2\%) on exact match for the Test-New set, which confirms an improved generalization capability by our translation model. However, for the Test-Repeated set we see a 8.5\% decrease in exact match with respect to the original approach (although we do beat our own implementation of the baseline by 25.9\% in this set, and by 18.4\% in the Test-New set).

% Table
\begin{table}[ht]
\label{tabla1}
\small
\centering
\begin{tabular}{|c|c|c|c|c|c|c|c|c|c|c}
\hline
\multirow{2}{*}{Architecture} & \multicolumn{5}{c|}{Test Repeated} & \multicolumn{5}{c|}{Test New} \\ \cline{2-11} 
 & F1 $\uparrow$ & M@0  $\uparrow$ & M@1 $\uparrow$ & M@2 $\uparrow$ & ED $\downarrow$ & F1 & M@0 & M@1 & M@2 & \multicolumn{1}{c|}{ED}  \\ \hline
Baseline (Zang) & \textbf{93.54} & \textbf{61.17} & \textbf{83.30} & \textbf{92.19} & \textbf{0.75} & 90.22 & 41.71 & 69.82 & 82.08 & \multicolumn{1}{c|}{1.22} \\ \hline
Baseline (Ours) & 91.67 & 44.43 & 76.93 & 89.16 & 1.01 & 90.89 & 43.41 & 72.64 & 87.25 & \multicolumn{1}{c|}{1.09} \\ \hline
Ours & 93.07 & 55.96 & 81.31 & 90.16 & 0.84 & \textbf{92.57} & \textbf{51.40} & \textbf{79.06} & \textbf{89.43} & \multicolumn{1}{c|}{\textbf{0.91}} \\ \hline
\end{tabular}
\caption{Results. The symbol $\uparrow$ indicates that higher results are better in the corresponding column; likewise $\downarrow$ indicates that lower is better.}
\end{table}

\paragraph{\textbf{Conclusions}} In this paper, we introduced multi-head attention as a useful mechanism for leveraging a knowledge base to improve natural language translations to a high-level behavioral language that is understandable and executable by robots, exhibiting a better performance on never-before-seen environments with respect to previous work. Future research efforts contemplate minimizing the lost performance over previously seen maps and doing a qualitative analysis of the resulting attention weights.

\paragraph{\textbf{Acknowledgments}} This work was partially funded by the Millennium Institute for Foundational Research on Data and Fondecyt grant 1181739.

\newpage
\bibliographystyle{acl}
\bibliography{references}

\begin{thebibliography}{}

\bibitem[\protect\citename{Chung \bgroup et al.\egroup }2014]{GRU}
Junyoung Chung, Çaglar G{\"u}lçehre, Kyunghyun Cho, and Yoshua Bengio.
\newblock 2014.
\newblock Empirical evaluation of gated recurrent neural networks on sequence
  modeling.
\newblock {\em ArXiv}, abs/1412.3555.

\bibitem[\protect\citename{Pennington \bgroup et al.\egroup }2014]{Glove}
Jeffrey Pennington, Richard Socher, and Christopher~D. Manning.
\newblock 2014.
\newblock Glove: Global vectors for word representation.
\newblock In {\em EMNLP}.

\bibitem[\protect\citename{Sepulveda \bgroup et al.\egroup
  }2018]{sepulveda2018deep}
Gabriel Sepulveda, Juan~Carlos Niebles, and Alvaro Soto.
\newblock 2018.
\newblock A deep learning based behavioral approach to indoor autonomous
  navigation.
\newblock In {\em 2018 IEEE International Conference on Robotics and Automation
  (ICRA)}, pages 4646--4653. IEEE.

\bibitem[\protect\citename{Tan and Bansal}2019]{tan2019lxmert}
Hao Tan and Mohit Bansal.
\newblock 2019.
\newblock Lxmert: Learning cross-modality encoder representations from
  transformers.
\newblock In {\em Proceedings of the 2019 Conference on Empirical Methods in
  Natural Language Processing and the 9th International Joint Conference on
  Natural Language Processing (EMNLP-IJCNLP)}, pages 5103--5114.

\bibitem[\protect\citename{Vaswani \bgroup et al.\egroup }2017]{vaswani}
Ashish Vaswani, Noam Shazeer, Niki Parmar, Jakob Uszkoreit, Llion Jones,
  Aidan~N. Gomez, Lukasz Kaiser, and Illia Polosukhin.
\newblock 2017.
\newblock Attention is all you need.
\newblock In {\em 31st Conference on Neural Information Processing Systems},
  NIPS '17.

\bibitem[\protect\citename{Zang \bgroup et al.\egroup
  }2018a]{zang-etal-2018-translating}
Xiaoxue Zang, Ashwini Pokle, Marynel V{\'a}zquez, Kevin Chen, Juan~Carlos
  Niebles, Alvaro Soto, and Silvio Savarese.
\newblock 2018a.
\newblock Translating navigation instructions in natural language to a
  high-level plan for behavioral robot navigation.
\newblock In {\em Proceedings of the 2018 Conference on Empirical Methods in
  Natural Language Processing}, pages 2657--2666, Brussels, Belgium,
  October-November. Association for Computational Linguistics.

\bibitem[\protect\citename{Zang \bgroup et al.\egroup
  }2018b]{Zang2018BehavioralIN}
Xiaoxue Zang, Marynel V{\'a}zquez, Juan~Carlos Niebles, Alvaro Soto, and Silvio
  Savarese.
\newblock 2018b.
\newblock Behavioral indoor navigation with natural language directions.
\newblock {\em Companion of the 2018 ACM/IEEE International Conference on
  Human-Robot Interaction}.

\bibitem[\protect\citename{Zhou \bgroup et al.\egroup
  }2020]{zhou2020recommending}
Yichao Zhou, Shaunak Mishra, Manisha Verma, Narayan Bhamidipati, and Wei Wang.
\newblock 2020.
\newblock Recommending themes for ad creative design via visual-linguistic
  representations.

\end{thebibliography}

\end{document}